\documentclass[letterpaper, 10 pt, conference]{ieeeconf}  

\usepackage[margin=1in]{geometry}
\usepackage{amsmath,amssymb}
\usepackage{atbegshi,picture}
\usepackage[pdftex]{graphicx} 
\usepackage{sidecap} 
\usepackage{tikz} 
\usepackage{pgfplots} 
\usepackage{float}
\usepackage{todonotes}
\usepackage{indentfirst}
\usepackage[linesnumbered,ruled]{algorithm2e}

\usetikzlibrary{arrows,shapes, arrows.meta}
\usetikzlibrary{decorations.markings}
\usetikzlibrary{decorations,decorations.pathmorphing}
\usepackage{url}
\usepackage[caption=false, font=footnotesize]{subfig}
\pgfplotsset{compat=1.7}

\pgfmathdeclarefunction{cn}{1}{%
  \pgfmathparse{(	(and(#1>=0,#1<.8) * ((2*(#1-.5))^2) ) +
  					(and(#1>=.8,#1<=1) * .75) )}%
}
\tikzstyle{cnstyle}=[domain=0:1, samples=100, ultra thick]
\pgfmathdeclarefunction{cy}{1}{%
  \pgfmathparse{(	(and(#1>=0,#1<1) * ((2*(#1-.5))^2) ) +
  					(and(#1>=1,#1<=1.2) * 1) )
  )}%
}
\tikzstyle{cystyle}=[domain=0:1.2, samples=100, ultra thick]

\pgfmathdeclarefunction{cg}{1}{%
  \pgfmathparse{(	(and(#1>=0,#1<=1) * atan(pi*#1)/90 ) )
  )}%
}
\tikzstyle{cgstyle}=[domain=0:1, samples=100, ultra thick]
\DeclareMathSizes{9}{9}{9}{9}

\usepackage{listings}
\usepackage{color}
\IEEEoverridecommandlockouts                              

\title{\LARGE \bf
Cost Adaptation for Robust Decentralized Swarm Behaviour
}

\author{Peter Henderson$^{1}$, Matthew Vertescher$^{2}$, David Meger$^{1}$, and Mark Coates$^{2}$
\thanks{$^{1}$Department of Computer Science,
        McGill University, Montreal, Quebec, Canada
        {\tt\small peter.henderson@mail.mcgill.ca}}%
\thanks{$^{2}$Department of Electrical, Computer, and Software Engineering,
        McGill University, Montreal, Quebec, Canada
        }%
}

\begin{document}

\maketitle
\thispagestyle{empty}
\pagestyle{empty}

\begin{abstract}
Decentralized receding horizon control (D-RHC) provides a mechanism for coordination in multi-agent settings without a centralized command center. However, combining a set of different goals, costs, and constraints to form an efficient optimization objective for D-RHC can be difficult. To allay this problem, we use a meta-learning process -- cost adaptation -- which generates the optimization objective for D-RHC to solve based on a set of human-generated priors (cost and constraint functions) and an auxiliary heuristic. We use this adaptive D-RHC method for control of mesh-networked swarm agents. This formulation allows a wide range of tasks to be encoded and can account for network delays, heterogeneous capabilities, and increasingly large swarms through the adaptation mechanism. We leverage the Unity3D game engine to build a simulator capable of introducing artificial networking failures and delays in the swarm. Using the simulator we validate our method on an example coordinated exploration task. We demonstrate that cost adaptation allows for more efficient and safer task completion under varying environment conditions and increasingly large swarm sizes. We release our simulator and code to the community for future work.
\end{abstract}

\section{INTRODUCTION}

In recent years, unmanned aerial vehicles (UAVs) have proven useful for a number of tasks -- such as searching, exploration, and area mapping~\cite{schwager2011eyes,banfi2017strategies,rekleitis2004limited}. However, the maximum flight time of these systems is still limited.
To accomplish such tasks efficiently, without need for a command-and-control center, decentralized swarm systems -- consisting of many agents which coordinate among themselves -- have been used. One method for coordination of a swarm is via decentralized receding horizon control (D-RHC)~\cite{keviczky2008decentralized}.
This involves formulating and optimizing a cost-based objective function using predictive information about neighbouring swarm members and the agent's goals. 

However, in a complex swarm, it is difficult to select the most efficient and safe combination of costs \emph{a priori}. Rather, agents receive feedback about their local environment and task performance during execution. Varying network delays, heterogeneous mixtures of robot capabilities, and other environmental factors may render a pre-defined D-RHC optimization objective ineffective. 
As such, we introduce the notion of cost adaptation in the context of a swarm D-RHC problem formulation. Using a heuristic-based guided search, we learn to generate the swarm's D-RHC objective by modifying and combining a set of human-generated prior cost and constraint functions.
We leverage the swarm's interaction with the environment to perform this meta-learning optimization, in our case through adaptive simulated annealing~\cite{ingber1996adaptive}.
This allows the adaptive agents to perform the present task optimized for safety and efficiency. We provide a simple method for combining goal-driven objectives and adapting to a myriad of environmental perturbations. This method has the added benefit of improving interpretability over a heuristic-optimized action policy (as with reinforcement learning in~\cite{hwangbo2017control}). The learned D-RHC objective (which uses an interpretable combination of human priors) can give insight into the decision-making process and can be reviewed/edited.

To implement and evaluate our adaptive D-RHC method, we focus on a cooperative exploration task with a UAV swarm, as this has been a widely cited problem with a variety of applications and goals~\cite{banfi2017strategies,rekleitis2004limited}. We first develop a set of costs and constraints based on natural flocking behaviour~\cite{Reynolds:1987:FHS:37402.37406} and an agent's goals, along with a decision-making framework for D-RHC.
For communication, we formulate a mesh network which can send messages within the swarm based on neighbourhood topology.
To determine an agent's current goals, we also implement a bidding system to actively divide and assign sub-goals within the swarm in a decentralized manner. This allows for large numbers of agents to extend the swarm and actively partition swarms if this results in more efficient global coordination. All of these goals (as in any cost-based formulation) can be difficult to balance, thus we augment the D-RHC decision-making framework with cost adaptation to learn the optimal D-RHC objective under varying conditions based on the aforementioned priors inspired from related D-RHC work. 

For evaluating adaptation to varying conditions, we construct a 3D simulation capable of mimicking packet loss and communication delays between agents. Through our simulator, we show the feasibility of our approach by accomplishing a cooperative exploration task. We demonstrate that the cost adaptation in our system allows for safe and efficient exploration even under large communication delays. The swarm can actively re-adapt to keep a safer distance between agents or even cluster together under negligible communication delays.
We release our simulator and code to the community\footnote{\url{https://github.com/Breakend/SocraticSwarm}}.


\section{Related Work}

Several coordinated swarm systems use centralized decision-making for coordination of complex tasks where fine-grained control is needed~\cite{preiss2017crazyswarm,shkurti2012multi}. However, as ours does, many others focus on decentralized coordination of UAVs for swarm task completion. In~\cite{vasarhelyi2014outdoor}, the authors present the ``first decentralized multicopter flock that performs stable autonomous outdoor flight''. A differential update rule for the velocity is derived, based on natural flocking rules and information belief about neighbouring swarm members. Other works favour continual cost optimization to find the next optimal velocity or position based on the belief of the neighbours states, rather than differential updates. This is accomplished by setting a time horizon and computing the optimal predicted action at the horizon. The action is taken, information is then gathered, and the optimization problem solved again repeatedly. These are called decentralized receding horizon control (D-RHC) methods~\cite{keviczky2008decentralized,keviczky2006decentralized}. Other works also present similar methods for cost-based swarm formulation for specific domains, including~\cite{schwager2011eyes}. In our work, we aim to move toward decentralized control by exchanging minimal coordination information using a mesh-network. This allows for extended swarms and dynamic swarm partitioning and re-joining in locales where it may not be feasible to establish a central command-and-control center. As such, we extend the D-RHC scheme posed in~\cite{keviczky2008decentralized}. 

Recent work has also used meta-learning to generate a loss function in deep learning settings for classification tasks~\cite{fan2018learning}. We add to this expanding research area and learn the D-RHC objective from a combination of human priors (pre-generated cost and constraint functions).

\section{Background and Preliminaries}





In Receding Horizon Control (RHC) control systems, a constrained optimization problem is used to determine the next action for a projected time frame ending with a ``receding horizon'' (i.e. the next timestep where an action should be computed)~\cite{mattingley2011receding}. RHC systems are used across many domains including decentralized swarm robotics~\cite{keviczky2006coordinated,keviczky2006decentralized}. A common theme throughout the RHC literature is its usage in mission critical systems where fast and reliable decision making is needed. With an optimized and appropriate set of cost functions, it has been shown that a system using RHC can ``perform near its physical limits"~\cite{mattingley2011receding} and, as such, is ideal for our application. In particular, we focus on D-RHC as a control scheme. D-RHC models can be generally formulated as solving a constrained optimization problem at each timestep. The D-RHC controller first predicts state information of its neighbouring nodes and any other variable data inputs at the next time horizon. To make these predictions it uses any model information it may have about the optimization functions of the other controllers or simply current state information like velocity. The system then solves the constrained cost minimization problem to determine the optimal solution at the next time horizon for a variable it can control, such as the velocity of the agent.



The ``boids'' model presented by Reynolds~\cite{Reynolds:1987:FHS:37402.37406} is based on natural flocking patterns in birds. Three core rules make up natural flocking. According to this work, each agent in a flock aims only to: ``steer to avoid crowding local flockmates" (separation), ``steer towards the average heading of local flockmates" (alignment), and ``steer to move toward the average position of local flockmates" (cohesion). By following these fundamental laws, simulated agents can mimic natural flocking behaviour often seen in birds. These have been used as inspiration for cost formulation in some cost-based methods~\cite{vasarhelyi2014outdoor,keviczky2008decentralized}, as we follow here.

\section{Method}



For the central decision-making problem, we frame our work in the context of a coordinated exploration task where the state space is easily divisible by an area grid. In this formulation, the mesh-networked decision-making process can be viewed in four parts: (1) knowledge distribution and gathering; (2) global goal assignment through bidding within agents connected through the mesh; (3) local action choice through cost-optimization; (4) meta-learning through cost adaptation. In the knowledge acquisition phase, agents broadcast their beliefs about their state to their neighbours. In the global goal assignment step, agents try to determine their next sub-goal via a decentralized bidding process. In the local action choice step, the information gained through propagation and bidding is used to optimize the local action choice used by low-level controllers via D-RHC. In the meta-learning step, the provided set of costs are used to formulate a new D-RHC objective to suit the properties of the environment that the agents are in. We decompose the global goal assignment from the D-RHC process to simplify the problem based on prior work~\cite{sheng2006distributed}. This also allows for more efficient task completion -- if the highest bidding areas are at separate ends of the space, the swarm can decompose itself into several swarms and rejoin later. A general outline of this can be seen in Algorithm~\ref{algo:general}.

\begin{algorithm}

    Divide task on-board each agent according to same process into a set possible goals $G_i$ associated with a cost $C_i$.
    
    \While{Task Not Complete}{
    
    \While{\textbf{not} WonBid}{
        bid = CalculateNextBid()
    
        SendToNeighbours(bid)
        
        ReceiveAndCorrectNeighbourBids()
    
        WonBid = CompleteBidding()
    
        \If{WonBid}{
           AddCost(bid.SubGoalCost)
        }
    }
    
    \While{\textbf{not} Completed(bid.SubGoal)}{
    belief=predictBeliefAt($t_{i+1}$)
    
    nextPos = optimize(costs, belief)
    
    applyVelocityToward(nextPos)
    
    broadcast(currentPosition)
    
    updates = ListenForPositionUpdates()
    
    updateBeliefState(updates)

    }
    }

    \caption{Agent Decision Process}
    \label{algo:general}
\end{algorithm}

\subsection{Knowledge Propagation and Mesh Networking}

We assume a fluid mesh network for our swarm: agents can drop in and out of the network without affecting the rest of the swarm (e.g. if that is the most optimal action to take or if the agent is destroyed). 
Each agent has the ability to send information to other swarm members about itself, the environment, or bidding results. Before an agent makes a decision, it must process all new messages and directly update its knowledge base. Each agent keeps a record of its current belief about a neighbour's position, goal, and the information about the world.

At the start of the decision loop, agents broadcast their position to their immediate neighbours in the mesh. Incoming updates are processed and placed into the knowledge base. Similar to D-RHC, agents will try to predict their neighbours positions at the next horizon using any position, velocity, and acceleration information in addition to the time elapsed since the message was sent (via timestamp on the information packet). For bidding, messages are propagated throughout the mesh and agents are made aware if new swarm members join.

\subsection{Bidding}
For global coordination, a task must be divided into sub-goals. In an exploration task, the sub-goals are formulated as searching an area subsection, or tile. To coordinate the distribution of these sub-goals, we formulate a consensus bidding mechanism wherein agents: calculate their next desired sub-goal to complete, broadcast a bid, and enter an auction process until they claim a sub-goal. All receiving agents keep a belief state of the state of all sub-goals (e.g. in bidding, claimed, completed). Bids are propagated through the mesh, unlike current agent positions which are only sent to immediate neighbours. Agents are not guaranteed to know the entire bidding process due to packet delays or drops and must make their decisions in a decentralized manner based on their current belief of the bidding process.

If an agent does not have a sub-goal assigned to it or a candidate sub-goal (where a bid has been submitted) it will generate a new bid. A bid is calculated as $g_j = w_\text{dist} D_j + w_\text{near} \lambda_{ij}$ based on~\cite{sheng2006distributed} and the largest bid from all unclaimed sub-goals is chosen. In the context of exploration, here $D_j$ is the distance to the centre of the tile and $\lambda_{ij}$ is the cohesion factor from other agents to the tile. This bid formulation encourages a choice of sub-goals that are close the current position, with some spread-out factor from other agents. This helps to reduce contention for tiles, which increases search efficiency. 

If an agent has the highest bid within some auction timeout, it claims the sub-goal. If there is a higher bid, it marks its own belief of the sub-goal as claimed by another agent and re-bids. Once an agent completes a sub-goal, it sends this update to its neighbours for propagation in the mesh as needed. With larger communication delays, race conditions can occur during bidding. To address this, we use a simple method: if an agent receives information contrary to its own beliefs, it attempts to correct its current belief state or send out corrective messages to the swarm. Some of these scenarios are described in Figure~\ref{fig:bidding}. We find that if we change bidding from propagation of messages through the mesh to the restricted sending of bids to immediate neighbours, the swarm is still able to complete tasks efficiently. This is because belief states are inherently updated through correction during the bidding process as we discuss earlier. These corrections cascade through the network, effectively accomplishing selective propagation. This effect is further seen when new members join a swarm or multiple swarms join together.



\tikzset{
    con/.style={draw=none,
                postaction={decoration={markings,mark=at position .8 with
                      {\pgftransformscale{3}\arrow[blue!50!black]{angle 90}}},decorate},
                postaction={draw,blue!50!black,decoration={snake,amplitude=1pt,
                      segment length=8pt},decorate}
               }
}

\tikzset{
    msg/.style={draw=none,
                postaction={decoration={markings,mark=at position .8 with
                      {\pgftransformscale{1}\arrow[black!50!black]{angle 90}}},decorate},
                postaction={draw,black!50!black,decoration={
                      segment length=4pt},decorate}
               }
}

\tikzstyle{agt} = [rectangle, draw, fill=orange!20,
    text width=3.2em, text centered, rounded corners, minimum height=1.5em]


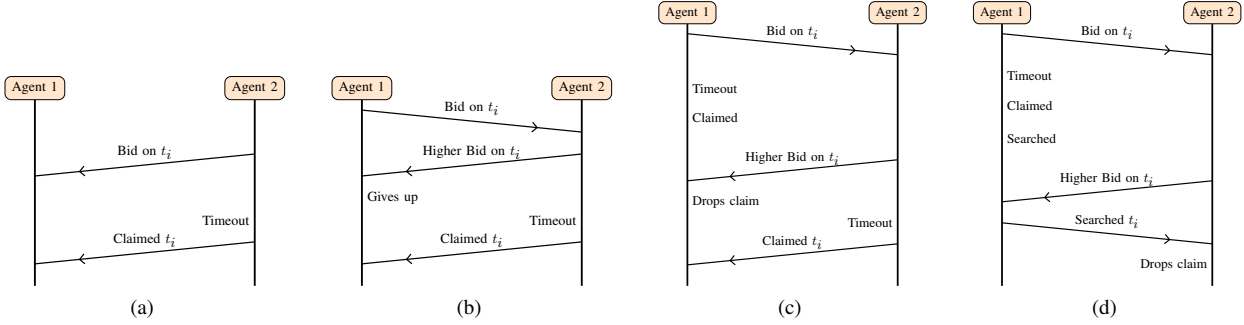
\begin{figure*} 
    \centering
  \subfloat[]{%
       \resizebox {.23\textwidth} {!} {
    \begin{tikzpicture}[node distance = 1cm, auto]
		\node [agt] (agt1) {\small Agent 1};
		\node [agt] (agt2) at ([shift={(5,0)}] agt1) {\small Agent 2};
		\draw [black, very thick] (agt1) -- (0,-4.5);
		\draw [black, very thick] (agt2) -- (5,-4.5);
		\draw [msg, thick] (5,-1.5) -> (0,-2) node[pos=0.5, anchor=south]{\small Bid on $t_i$};
		\node[anchor=east] (tout) at (5,-3) {\small Timeout};
		\draw [msg, thick] (5,-3.5) -> (0,-4) node[pos=0.5, anchor=south]{\small Claimed $t_i$};
	\end{tikzpicture}}}
    \label{1a}\hfill
  \subfloat[]{%
        \resizebox {.23\textwidth} {!} {
    \begin{tikzpicture}[node distance = 1cm, auto]
		\node [agt] (agt1) {\small Agent 1};
		\node [agt] (agt2) at ([shift={(5,0)}] agt1) {\small Agent 2};
		\draw [black, very thick] (agt1) -- (0,-4.5);
		\draw [black, very thick] (agt2) -- (5,-4.5);
		\draw [msg, thick] (0,-0.5) -> (5,-1) node[pos=0.5, anchor=south]{\small Bid on $t_i$};
		\draw [msg, thick] (5,-1.5) -> (0,-2) node[pos=0.5, anchor=south]{\small Higher Bid on $t_i$};
		\node[anchor=west] (givu) at (0,-2.5) {\small Gives up};
		\node[anchor=east] (tout) at (5,-3) {\small Timeout};
		\draw [msg, thick] (5,-3.5) -> (0,-4) node[pos=0.5, anchor=south]{\small Claimed $t_i$};
	\end{tikzpicture}}}
    \label{1b}\hfill
  \subfloat[]{%
        \resizebox {.22\textwidth} {!} {
    \begin{tikzpicture}[node distance = 1cm, auto]
		\node [agt] (agt1) {\small Agent 1};
		\node [agt] (agt2) at ([shift={(5,0)}] agt1) {\small Agent 2};
		\draw [black, very thick] (agt1) -- (0,-6.5);
		\draw [black, very thick] (agt2) -- (5,-6.5);
		\draw [msg, thick] (0,-0.5) -> (5,-1) node[pos=0.5, anchor=south]{\small Bid on $t_i$};
		\node[anchor=west] (givu) at (0,-1.8) {\small Timeout};
		\node[anchor=west] (givu) at (0,-2.5) {\small Claimed};
		\draw [msg, thick] (5,-3.5) -> (0,-4.0) node[pos=0.5, anchor=south]{\small Higher Bid on $t_i$};
		\node[anchor=west] (tout) at (0,-4.5) {\small Drops claim};
		\node[anchor=east] (givu) at (5,-5.0) {\small Timeout};
		\draw [msg, thick] (5,-5.5) -> (0,-6) node[pos=0.5, anchor=south]{\small Claimed $t_i$};
	\end{tikzpicture}}}
    \label{1c}\hfill
  \subfloat[]{%
        \resizebox {.22\textwidth} {!} {
    \begin{tikzpicture}[node distance = 1cm, auto]
		\node [agt] (agt1) {\small Agent 1};
		\node [agt] (agt2) at ([shift={(5,0)}] agt1) {\small Agent 2};
		\draw [black, very thick] (agt1) -- (0,-6.5);
		\draw [black, very thick] (agt2) -- (5,-6.5);
		\draw [msg, thick] (0,-0.5) -> (5,-1) node[pos=0.5, anchor=south]{\small Bid on $t_i$};
		\node[anchor=west] (givu) at (0,-1.5) {\small Timeout};
		\node[anchor=west] (givu) at (0,-2.2) {\small Claimed};
		\node[anchor=west] (givu) at (0,-3.0) {\small Searched};
		\draw [msg, thick] (5,-4.0) -> (0,-4.5) node[pos=0.5, anchor=south]{\small Higher Bid on $t_i$};
		\draw [msg, thick] (0,-5.0) -> (5,-5.5) node[pos=0.5, anchor=south]{\small Searched $t_i$};
		\node[anchor=east] (tout) at (5,-6.0) {\small Drops claim};
	\end{tikzpicture}}}
     \label{1d} 
  \caption{Illustrations of several bidding scenarios. (a) shows the common no contention bid. (b) shows two agents bidding over the same tile where a winner is achieved without race conditions. (c) illustrates an agent bidding on an already claimed tile. The agent with the lower bid drops its claim. (d) shows an agent claiming a searched tile. The agent that searched the tile corrects its neighbours' knowledge.}
  \label{fig:bidding} 
\end{figure*}

\subsection{Local Action Choice through D-RHC}

As aforementioned, D-RHC optimizes for the best possible position to be at the next time horizon given the cost function. While we rely on cost adaptation to provide the optimal cost function, we generate a set of initial priors for the guided search to modify and combine. We formulate our costs with inspiration of previous work, aiming for finite, normalized, and interpretable cost functions. We highlight, however, that given any set of reasonable costs and constraints the cost adaptation method should converge to a functioning objective. 

We define three costs (priors) which the adaptation method can used to generate the D-RHC objective: cohesion, safety, and goal. These are based on the formulations seen in~\cite{vasarhelyi2014outdoor,keviczky2008decentralized}. While we initially considered the rote boids formulation (i.e. alignment, separation, and cohesion), we empirically found that we could combine separation and cohesion into one term and remove the alignment penalty for simplicity and reduction of the adaptation search space. We further modify the costs such that they are normalized between $0$ and $1$. This ensures that the cost adaptation method does not need to re-scale significantly over time based on exploding costs. \\

\noindent \textbf{Cohesion}
Our combined cohesion cost is:

\begin{gather} \label{eq:cost-cohesion}
f(d_j, r_c) = \begin{cases}
  	  (2d_j/r_c) - 1 & \text{if } d_j < 0.75r_c \\
  	  C_{penalty} & \text{otherwise }
  \end{cases} \\
c_{\eta} = \min \left( 1, \frac{\sum_{j = 1}^{N}{\alpha^j f(d_j, r_c)} }{N} \right)
\end{gather}

Where $d_j = \text{dist}(p_j, p_{i})$ and $d_1 \leq d_2 \dots \leq d_{N}$ (i.e. the distance are sorted in order). Here $N$ designates the local neighbour space (i.e. the number of immediately connected neighbours in the mesh). $r_c$ is the communication range; $\alpha \in [0,1]$ is a fading factor. \\

\noindent \textbf{Safety Cost}
While the safety cost can adapt to many contexts, here we simply formulate it as keeping a safe altitude off of the ground. Note, that we explicitly model collision with other agents as a constraint, while altitude is a cost. This is by design, as in the case of landing, the altitude safety cost can be decreased.

\begin{gather}
\label{eq:cost-altitude}
 c_{z} = \min \left( 1,
  \begin{cases}
  	((z_{i} / z_{\text{min}}) - 1)^2 & \text{if } z_{\text{min}} > z_{i} \\
  	((z_{i} -  z_{\text{min}}) / z_{\text{max}})^2 & \text{otherwise }
  \end{cases}
  \right)
\end{gather}
\\
Here, $z_i$ is the altitude to be tested and $z_{\min}, z_{\max}$ are the bounds for the altitude.\\

\noindent \textbf{Goal Cost}
Lastly, we must formulate a goal cost associated with the completion of a sub-goal. Once a sub-goal has been claimed, this goal cost is added to the D-RHC optimization problem, driving task completion in local actions. This is formulated as:

\begin{equation} \label{eq:cost-goal}
c_{g} = \min \left( 1, \frac{2}{\pi} \arctan \left( \frac{\text{dist}(p_i,p_{\text{goal}})}{C_{\text{dist}}} \right) \right)
\end{equation}

Where $p_i$ is the agent's position to be tested and $p_{\text{goal}}$ is the position of the centre of the goal tile which the agent intends to search. $C_{\text{dist}}$ is a centroid point which determines the slope of the cost according to $\arctan$.\\

\noindent \textbf{Full Optimization Problem}
Thus our modifiable swarm action decision process can be viewed as a weighted sum of costs resulting in the optimization problem:

\begin{gather} \label{eq:nextpoint}
\begin{matrix}
\min & ( w_\eta c_{\eta}
	+ w_\text{z} c_{z}
	+ w_\text{g} c_{g} )\\
\text{s.t.} & p^{t+1}_{i} < p^t_{i} + (v^{t}_{i} + a^{t}_{i} \delta t) \delta t\\
& \forall_{j \in \text{neighbours}} \text{ dist}(p^{t+1}_i, p^{t+1}_j) > \Delta_\text{min} \\
\end{matrix}
\end{gather}

Here, $w_\eta, w_y, w_g$ are the cost weights. $v^{t}_{i}$ is the current velocity, $a^{t}_{i}$ is the current acceleration, $\delta t$ is the time until the next horizon, $p^{t+1}_{i}$ is the next position with cost to be minimized, $p^t_{i}$ is the current position, $p^{t+1}_{j}$ is a neighbour's predicted position at the next horizon, and $\Delta_{\min}$ is some minimum distance kept between an agent and any neighbour.

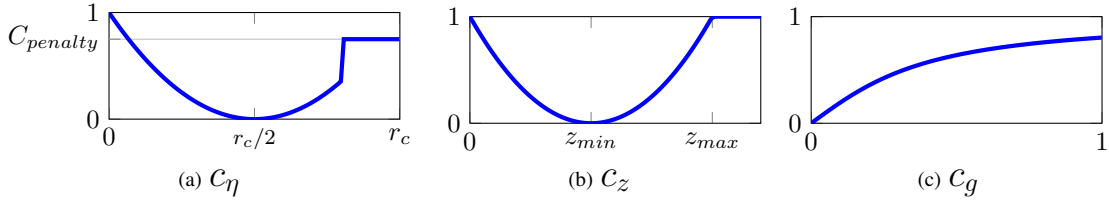
\begin{figure*}
\centering
\subfloat[\Large$c_\eta$] {
\begin{tikzpicture}
  \begin{axis}[width=.33\textwidth,height=3.0cm,xlabel={},ylabel={},xmin=0,xmax=1,ymin=0,
ymax=1,xtick=\empty, ytick=\empty, xtick={0,.5,1}, xticklabels={$0$,\footnotesize  $r_c/2$,$r_c$}, ytick={0,1}, yticklabels={$0$,$1$},extra y ticks={.75},extra y tick style={grid=minor, yticklabel={$C_{penalty}$}, yticklabel pos=left, yticklabel style={anchor=east}}]  
    \addplot[cnstyle, blue]{cn(x)}; 
  \end{axis}
\end{tikzpicture}
}
\subfloat[\Large$c_z$] {
\begin{tikzpicture}
  \begin{axis}[width=.33\textwidth,height=3.0cm,xlabel={},ylabel={},xmin=0,xmax=1.2,ymin=0,
ymax=1,xtick=\empty, ytick=\empty, xtick={0,.5,1}, xticklabels={$0$,$z_{min}$,$z_{max}$}, ytick={0,1}, yticklabels={$0$,$1$}]  
    \addplot[cystyle, blue]{cy(x)}; 
  \end{axis}
\end{tikzpicture}
}
\subfloat[\Large$c_g$] {
\begin{tikzpicture}
  \begin{axis}[width=.33\textwidth,height=3.0cm,xlabel={},ylabel={},xmin=0,xmax=1,ymin=0,ymax=1,xtick=\empty, ytick=\empty, xtick={0,1}, xticklabels={$0$,$1$}, ytick={0,1}, yticklabels={$0$,$1$}]  
    \addplot[cgstyle, blue]{cg(x)}; 
  \end{axis}
\end{tikzpicture}
}
\caption{Visualizations of each of the three cost functions. $c_\eta$, $c_z$ and $c_g$ describe the cohesion, safety and goal cost respectively.}
\label{fig:costviz}
\end{figure*}

Figure~\ref{fig:costviz} shows the functions modelled by these costs. 
Note, that we formulate collision as a constraint, bounding the next optimal position such that it is not within some sphere of radius $\Delta_\text{min}$ surrounding an agent. For higher communication delays, this $\Delta_\text{min}$ can be adapted to prevent agents from risking collision with an uncertain agents position. Furthermore, we bound the search space to only points that are reachable by the agent by the next horizon based on the agent's current velocity and acceleration.

To minimize our D-RHC objective, we attempted several different techniques. We investigated Particle Swarm Optimization with constraints and Differential Evolution with constraints\footnote{Both PSO and DE from C\# framework found at:\\ \url{http://www.hvass-labs.org/projects/swarmops/cs/}}. Both PSO and DE performed well, but DE with constraints proved to be the least computationally intensive approach with acceptable solutions. As such, we simply use this method as a black box for constrained optimization of D-RHC problem.


\section{Cost Adaptation}
Finally, we introduce a meta-learning process -- cost adaptation -- which uses the human generated prior costs (defined previously) to generate a new D-RHC objective. Cost adaptation optimizes an auxiliary heuristic in successive trials based on swarm performance under the current system dynamics (such as conditions with heavy packet loss) through a guided search mechanism. In the previously defined set of human priors (the costs and constraints), the variables in the meta-learning search space are: $w_{\eta}, w_y, w_g, \Delta_{\min}, C_{penalty} \in \Theta$. 
Each variable in the cost adaptation problem is bounded such that $\theta = [\Theta_{\min}, \Theta_{\max}]$, where all are bounded between $0$ and $1$, except for $\Delta_{\min}$, which has a large bound of $100$. We evaluate each trial using a set of costs and constraints according to a heuristic, $E_{c}$. We generate this heuristic based on our prior intuition, just as a reward is generated in many reinforcement learning domains~\cite{sutton1998reinforcement}. Generally, we try to normalize each part of the heuristic function between $0$ and $1$, re-weighting portions by importance. We give penalties for collisions between agents and bonuses for completing tasks quickly and completely.


\begin{equation}
E_{c} = c_{\text{time}} + 2c_{\text{nsrh}} + 4c_{\text{clsn}}
\end{equation}
\begin{equation}
c_{\text{time}} = \frac{t_{\text{trial}}}{t_{\text{max}}}
\end{equation}
\begin{equation}
c_{\text{clsn}} = \begin{cases}
0.25 + 0.75\frac{n_{\text{clsn}}}{n_{\text{agents}}} &, n_{\text{clsn}} > 0\\
0 &, \text{otherwise}
\end{cases}
\end{equation}

Here, $p_{complete}$ is the fraction of the task completed (for the coordinated exploration task this is the percentage of tiles searched); $n_{\text{clsn}}$ is the number of agent collisions with the total number of agents being: $n_{\text{agents}}$; $t_{\text{trial}}$ is the time it took to complete a trial and $t_{\text{max}}$ is the maximum allowed trial time.

While we initially formulated the cost adaptation problem with simulated annealing~\cite{van1987simulated}, we found this to be a sample inefficient search method. Since the desire is to find a suitable D-RHC objective in as few simulations as possible, we turned to Adaptive Simulated Annealing (ASA)~\cite{ingber1996adaptive, ingber2012adaptive}. We found poor convergence properties with the rote version of ASA, as such, we modify it for our own purposes, seen in Algorithm~\ref{algo:costreweight}. We choose a temperature decay of $.95$ and a re-annealing schedule of $25$ according experimental search performance.

\begin{algorithm}
    \SetKwInOut{Input}{Input}
    \SetKwInOut{Output}{Output}

    \Input{$w \in W$ for all weights assigned to costs and constraints. $w_{\min}, w_{\max}, \forall w \in W$}
    \Output{$\bar{w} \in \bar{W}$}
     Set a temperature $T_C = T_0 = .95^{-\text{MAX\_TRIALS}}$
     
     Initialize $\hat{w_0} = \bar{w} = w  ; \forall w \in W, \hat{w} \in \hat{W}, \bar{w} \in \bar{W}$
     
     Set $E_{\min} = \infty$
     
    \For{$i=0,1,2,\dots$,\text{MAX\_TRIALS}}
    {
    Sample $\hat{w}_{i+1} \in \hat{W}_{i+1}$ with the random variable $Y_i \sim U[{(\bar{w}_{\min} - \bar{w})}\frac{T_C}{T_0}, {(\bar{w}_{\max} - \bar{w}_i)}\frac{T_C}{T_0}]$ according to:
    $$\hat{w}_{i+1} = w_i + y_i; w_{i+1} \in W_{i+1}, w_i \in W_{i}$$
    Run trial with $\hat{W}$ and evaluate heuristic $E_{c}$.
    
    \If{$E_{c} < E_{\min}$ \textbf{or} $e^\frac{E_{c_{prev} - E_{c}}}{T_C} > v \sim U[0,1]$}
    {
    $E_{\min}  = E_{c}$\newline
    Set $\bar{w} = \hat{w_i}; \forall \hat{w_i} \in \hat{W_i}, \bar{w} \in \bar{W}$
    }
    
    Set $T_C = T_0 (.95)^i$
    
    \If{$i\mod 25 = 0$}{
        Re-anneal, by setting $T_C = T_0$
    }
    
    }
    \caption{Adapting Costs and Constraints}
    \label{algo:costreweight}
\end{algorithm}

\tikzstyle{learning} = [rectangle, draw, fill=green!20,
    text width=3em, text centered, rounded corners, minimum height=1.5em]
\tikzstyle{world} = [rectangle, draw, fill=blue!20,
    text width=3em, text centered, rounded corners, minimum height=1.5em]
\tikzstyle{agent} = [rectangle, draw, fill=orange!20,
    text width=3.2em, text centered, rounded corners, minimum height=1.5em]
\tikzstyle{module} = [rectangle, draw, fill=orange!5,
    text width=4.2em, rounded corners, minimum height=1.5em]
\tikzstyle{line} = [draw, -latex']

\tikzset{
  owarrow/.style={
    decoration={markings,mark=at position 1 with {\arrow[scale=8,black]{>}}},
    postaction={decorate},
    shorten >=0.4pt}}

\tikzset{
  twarrow/.style={
    decoration={markings,mark=at position 0 with {\arrow[scale=4,black]{<}}, mark=at position 1 with {\arrow[scale=4,black]{>}}},
    postaction={decorate},
    shorten >=0.4pt}}

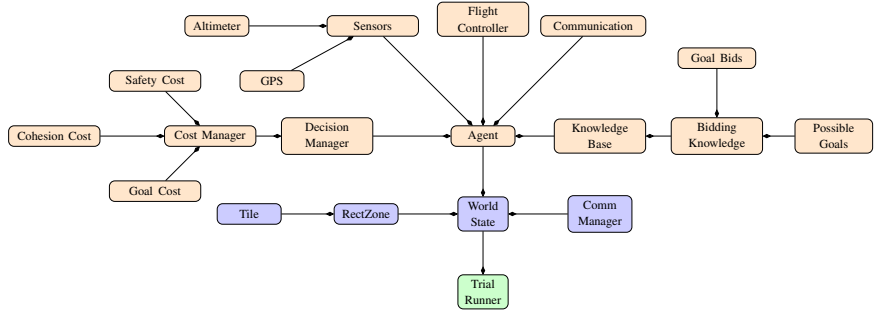
\begin{figure*}[!htb]
\centering
\resizebox{.7\textwidth}{!}{
\begin{tikzpicture}[node distance = .5cm, auto]
\node [learning] (tr) {\small Trial Runner};
\node [world, above of=tr, node distance=2cm] (ws) {\small World State};
\node [world, text width=4em, right of=ws, node distance=3cm] (cm) {\small Comm Manager};

\node [agent, text width=4em, above of=ws, node distance=2cm] (agt) {\small Agent};
\node [agent, text width=7em, above right of=agt, node distance=4cm] (comm) {\small Communication};

\node [agent, text width=6em, right of=agt, node distance=3cm] (kb) {\small Knowledge Base};
\node [agent, text width=6em, right of=kb, node distance=3cm] (bk) {\small Bidding Knowledge};
\node [agent, text width=5em, right of=bk, node distance=3cm] (st) {\small Possible Goals};
\node [agent, text width=5em, above of=bk, node distance=2cm] (tb) {\small Goal Bids};

\node [world, text width=4em, left of=ws, node distance=3cm] (rz) {\small RectZone};
\node [world, text width=4em, left of=rz, node distance=3cm] (tile) {\small Tile};

\node [agent, text width=6em, above left of=agt, node distance=4cm] (sen) {\small Sensors};
\node [agent, text width=4em, left of=sen, node distance=4cm] (alt) {\small Altimeter};
\node [agent, text width=4em, below right of=alt, node distance=2cm] (gps) {\small GPS};

\node [agent, text width=6em, left of=agt, node distance=4cm] (dm) {\small Decision Manager};
\node [agent, text width=6em, left of=dm, node distance=3cm] (costm) {\small Cost Manager};
\node [agent, text width=6em, left of=costm, node distance=4cm] (nearc) {\small Cohesion Cost};
\node [agent, text width=6em, above left of=costm, node distance=2cm] (altc) {\small Safety Cost};
\node [agent, text width=6em, below left of=costm, node distance=2cm] (goalc) {\small Goal Cost};

\node [agent, text width=6em, above of=agt, node distance=3cm] (flight) {\small Flight Controller};

\draw [-{Diamond}] (ws) -- (tr);
\draw [-{Diamond}] (tile) -- (rz);
\draw [-{Diamond}] (rz) -- (ws);
\draw [-{Diamond}] (cm) -- (ws);
\draw [-{Diamond}] (gps) -- (sen);
\draw [-{Diamond}] (alt) -- (sen);

\draw [-{Diamond}] (agt) -- (ws);
\draw [-{Diamond}] (sen) -- (agt);
\draw [-{Diamond}] (comm) -- (agt);
\draw [-{Diamond}] (kb) -- (agt);
\draw [-{Diamond}] (dm) -- (agt);
\draw [-{Diamond}] (flight) -- (agt);

\draw [-{Diamond}] (bk) -- (kb);
\draw [-{Diamond}] (st) -- (bk);
\draw [-{Diamond}] (tb) -- (bk);

\draw [-{Diamond}] (costm) -- (dm);
\draw [-{Diamond}] (nearc) -- (costm);
\draw [-{Diamond}] (altc) -- (costm);
\draw [-{Diamond}] (goalc) -- (costm);

%
%



\end{tikzpicture}}
\caption{Simplified class diagram of the simulation structure. The colours green, blue and orange represent the Learning, World State and Agent modules respectively. The Learning module consists of the trial runner, responsible for conducting a single trial. The World State module facilitates inter-agent communication and maintains the true state of the task. It also serves as the communication manager. The Agent module consists of several sub modules designed to support each of the required functionality from movement to decision making. Agents communicate with each other through the world module.}
\label{fig:simulator-overview}
\end{figure*}

\section{Experiments}

To validate the benefits of cost adaptation, we formulate several sets of experiments where we use a default setting (optimized for the base problem with no variation) and compare it against cost adaptation under various environment perturbations. In these experiments, we use an example coordinated exploration problem. Here, the task is to search a given a $600 m \times 400 m$ area. This area can be divided into 384 ($25m \times 25 m$) tiles. To search a tile properly, an agent must be a distance of 5 meters or less from the centre of the tile and 40 meters above the ground. This is to simulate taking an aerial photograph. The goal is to search the entire space in a decentralized manner in the least amount of time possible without collisions. The timeout for tiles after the first bid (auction time, $t_{\text{auction}}$) was 0.5 seconds. Agents update their next desired location every $t_{\text{update}}=0.05\text{s}$ and send agent update message every $t_{\text{broadcast}}=0.25\text{s}$. The communication range was set to $200 m$ based on several known 802.11s capable chipsets~\cite{li2013routing}. The maximum velocity of an agent was set to $40 ms^{-1}$ in any direction. With maximum vertical acceleration at $6 ms^{-2}$ and horizontal acceleration at $3 ms^{-2}$. These settings allow for a balance between realistic dynamics and speed of simulation.\\

\noindent\textbf{Simulator} Several promising 3D multi-agent robotics simulators exist, including Stage~\cite{vaughan2008massively} and Argos~\cite{pinciroli2012argos}. However, we found that existing simulators provided a large amount of overhead in fast implementation of mesh-networked agents or were poorly maintained. Instead, we built our simulation on the actively maintained Unity3D game engine\footnote{\url{https://unity3d.com}}. Unity supports basic physics simulations, networking, and all other necessary components for our swarm simulation including the ability to add visual cues which indicate algorithm performance.

The overall architecture of the simulator encompasses three main modules: Learning, World State, and Agent. The Learning module facilitates the execution of multiple trials for weight optimization and cost adaptation. The World State module is in charge of running a single trial, simulated communication, and keeping track of the overall world state. Finally, the Agent module consists of several sub modules designed to simulate a swarm agent and decision making algorithms. A simplified class diagram of the framework can be seen in Figure \ref{fig:simulator-overview}.

\begin{figure*}[ht!]
\centering
\includegraphics[width=.22\textwidth]{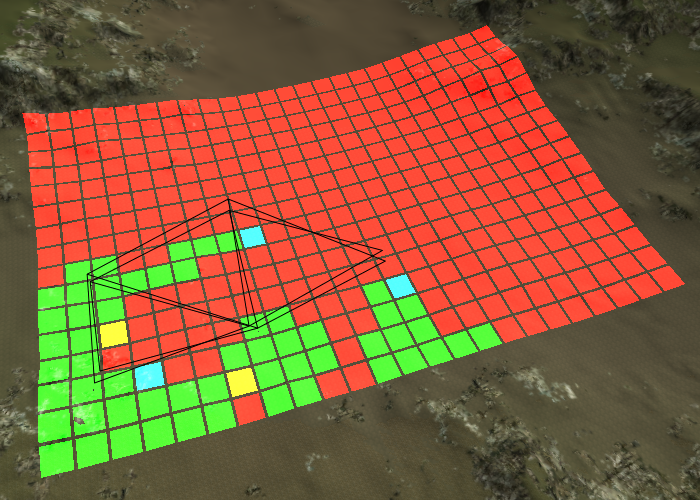}
 \includegraphics[width=.22\textwidth]{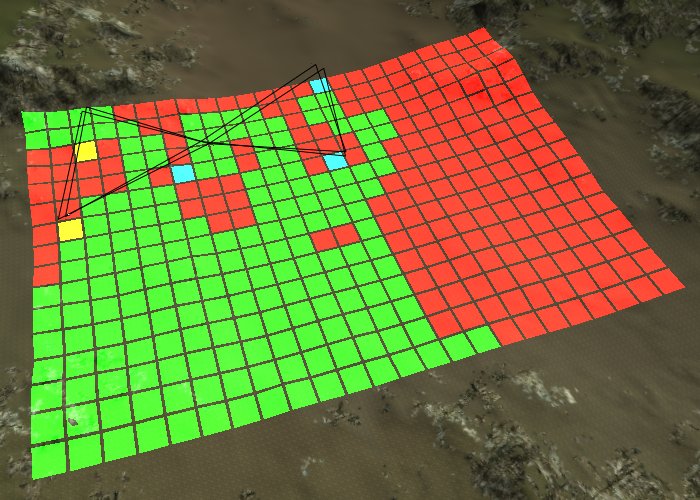} 
\includegraphics[width=.22\textwidth]{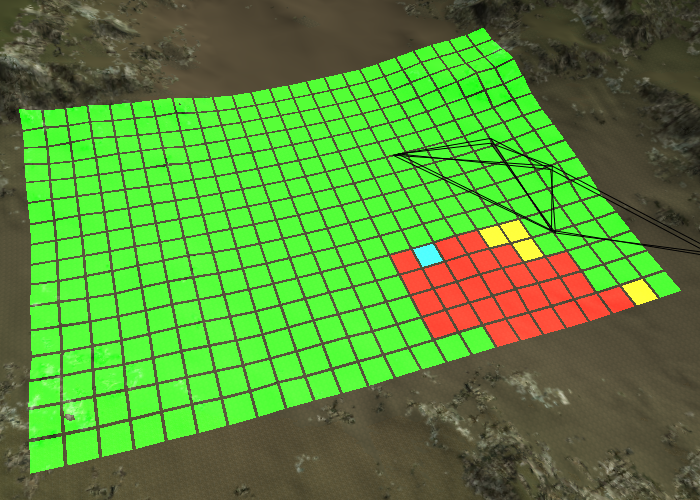} 
 \includegraphics[width=.22\textwidth]{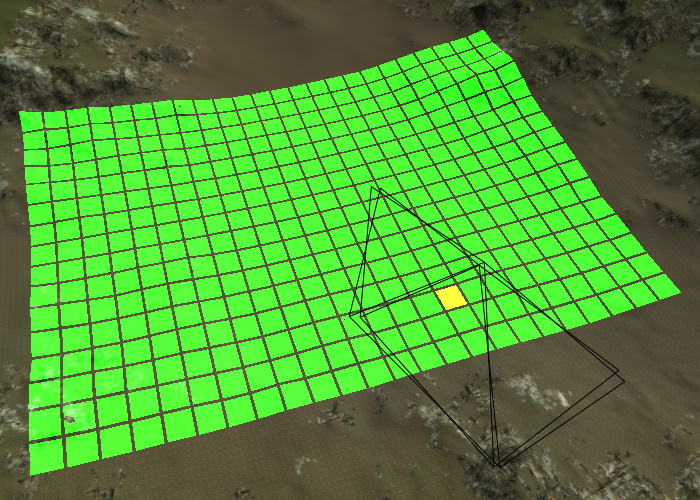} 
\caption{Five agents searching 384, $25 \times 25 m^2$ tiles. Agent start in the bottom left corner and search clockwise around the grid as illustrated by the spread of the green tiles. Black lines show an agents position to where it believes each of the other agents to be. Agents tend to search tiles at a comfortable distance to swarm. Red, blue, yellow, green are the states of the tiles (not claimed, in auction, claimed, searched).}
\label{ref:results-images}
\end{figure*}

Each agent has a simulated GPS sensor for position and altitude as well as an IMU for heading, velocity, acceleration. To simulate the real world inaccuracy of sensors, random noise of up to two meters is added to the agents' position. We also add an simulated altimeter based on ray-casting to the ``sea-level''. A simple flight controller is implemented which calculates the necessary pitch, roll, yaw velocities to move toward the next desired position. The maximum velocity and acceleration are scaled in the positive and negative y directions representing the agility of the quad copter to gain or drop altitude quickly. Agents send either position updates or bidding related messages through the WorldStateManager, which represents the mesh network protocol's lower communication layers. Figure \ref{ref:results-images} shows an example of five agents in simulation\footnote{Video demo: \url{https://youtu.be/2TUSXMo493I}}.


\section{Results}

\subsection{Adaptation to Network Delays}
Table~\ref{tab:results-comm} compares using increasing communication delays with five agents. The average completion time ($\mu_t$) and variance ($\sigma_t^2$) across 100 simulated trials -- along with the percent of area searched in the restricted time frame (300 seconds) and number of collisions between agents in that time frame. Communication delay was variable with the average delay as shown in the table. The control scenario consists of each agent having a pre-calculated flight plan which avoids all other agents. As the communication delay increases past the auction time of $0.5$ seconds, collisions and slower search times appear when the cost and nearness constraint is not properly tuned for uncertainty in neighbouring position.

In the second half of Table~\ref{tab:results-comm}, costs were adapted with ASA for 50 iterations, then 100 trials were used for evaluation. As can be seen, performance drastically improves as the D-RHC objective is adapted to complete tasks in the safest and most efficient manner possible. While we note that the performance of the control in this scenario is obviously faster, it is not robust to changing goals and conditions (such as destruction of members of the swarm). In our approach, the swarm recovers and picks up other workers' tasks if they are not completed.

\begin{table}
\centering 
\caption{Communication delay trials}
\label{tab:results-comm}
\footnotesize{
    \begin{tabular}{ | c | c | c | c | c |}
    \hline
    $T_{\text{delay}}$ (s) & $\mu_{t}$ (s) & $\sigma_{t}^2$ (s) & $\mu_{\text{\% Searched}}$ & $\mu_{\text{collisions}}$ \\ \hline
        \multicolumn{5}{|c|}{Without Adaptation}\\
 \hline
    Centralized Flight Plan & 107.08 & 0.47   & 100 & 0   \\ \hline 
	0       & 175.15 & 48.26  & 100 & 0   \\ \hline
	0.2     & 177.30 & 42.79  & 100 & 0   \\ \hline
	0.4     & 174.68 & 53.80  & 100 & 0   \\ \hline
	0.8     & 171.60 & 49.73  & 100 & 0   \\ \hline
	1.6     & 198.86 & 370.13 & 100 & 0.12   \\ \hline
	3.2     & 221.56 & 449.73 & 98  & 2.25 \\ \hline
    \multicolumn{5}{|c|}{With Adaptation}\\
    \hline
	  0       & 164.44 & 53.86 & 100 & 0   \\ \hline
	  0.2     & 162.22 & 45.12 & 100 & 0   \\ \hline
  	0.4     & 168.15 & 56.72 & 100 & 0   \\ \hline
  	0.8     & 171.34 & 33.14  & 100 & 0   \\ \hline
  	1.6     & 195.93 & 67.26 & 100 & 0   \\ \hline
  	3.2     & 191.07 & 77.45 & 100 & 0 \\ \hline
    \end{tabular}}

\vspace{-1.5em}
\end{table}



\subsection{Adaptation to Heterogenous Capabilities}
To further expand on our analysis of the system's ability to adapt to new conditions, we investigate using heterogeneous mixtures of agents. For these purposes, we spawn 10 agents and limit trials to 100 seconds such that goal is to complete as many tiles as possible in this time with heterogeneous dynamics. We use the same 384 tile grid and run 100 trials. We vary the maximum velocity of the agents by adding a fixed Gaussian noise of the form $Z \sim \mathcal{N}(0,\sigma_{\text{maxVel}})$ to the maximum velocity of each agent at the start of the experiment such that every agent has different flight dynamics. 

\begin{table}
\centering
\caption{Adaptation to velocity Gaussian noise}
\label{tab:results-vel}
    \footnotesize{\begin{tabular}{ | c | c | c | c |}
    \hline
    $\sigma_{maxVelocity}$& $\mu_{\text{\% Searched}}$ & $\sigma_{\text{\% Searched}}^2$ & $\mu_{\text{collisions}}$ \\ \hline
    \multicolumn{4}{|c|}{Without Adaptation}\\ \hline
	  10       & 51.4 \% & 3.9 \% & 0   \\ \hline
	  20     & 30.7 \% & 7.4\% & 0  \\ \hline
  	  30     & 40.1 \%& 5.7\% & 0 \\ \hline
    \multicolumn{4}{|c|}{With Adaptation}\\
    \hline
	  10      & 63.2 \% & 3.4 \% & 0  \\ \hline
	  20     & 53.2 \% & 7.9\% &0   \\ \hline
  	  30     & 70.1 \% & 13.5\% &0  \\ \hline  	  
    \end{tabular}}

\end{table}

\begin{table}
\centering
\caption{Adaptation to acceleration Gaussian noise}
\label{tab:results-acc}
    \footnotesize{\begin{tabular}{ | c | c | c | c |}
    \hline
    $\sigma_{acceleration}$ & $\mu_{\text{\% Searched}}$ & $\sigma_{\text{\% Searched}}^2$ & $\mu_{\text{collisions}}$ \\ \hline
    \multicolumn{4}{|c|}{Without Adaptation}\\ \hline
	  .5       & 51.4 \% &  1.8\% & 0   \\ \hline
	  1.0     &  61.1\% &  3.3\% & 0  \\ \hline
  	  2.0     &  67.2\%&  2.4\% & 0 \\ \hline
  	  2.5     &  51\%&  12.15\% & 0 \\ \hline
    \multicolumn{4}{|c|}{With Adaptation}\\
    \hline
	  .5      &  63.3\% &  0.2\% & 0  \\ \hline
	  1.0     &  70.1\% &  0.8\& &0   \\ \hline
  	  2.0     &  73.6\% &  3.6\% &0  \\ \hline  	
  	  2.5     &  77.9\%&  6.7\% & 0 \\ \hline
    \end{tabular}}


\end{table}

We use the original parameters used for the results seen Table~\ref{tab:results-vel} and perform 50 iterations of ASA to re-adapt our D-RHC cost formulation before our 100 trial iterations. We similarly add a fixed Gaussian noise of the form $Z \sim \mathcal{N}(0,\sigma_{\text{acc}})$ to the acceleration of each agent in another set of experiments. As demonstrated in Tables~\ref{tab:results-vel} and~\ref{tab:results-acc}, we find that in nearly every case, adaptation significantly improves performance. At higher variances in heterogeneity, we found that the non-adaptive swarm sometimes tried to to stick with a slower agent and ended up hurting efficiency. In these cases, the adaptive swarm found converged to a D-RHC objective which left behind lagging members by lowering the cohesion cost.

\begin{table}
\centering 
\caption{Trials on scalability to number of agents. }
\label{tab:results-agents}
\footnotesize{\begin{tabular}{ | c | c | c | c | c |} 
    \hline
    Agents & $\mu_{t}$  (s) & $\sigma_{t}^2$ (s) & $\mu_{\text{collisions}}$ \\ \hline \hline
	5 	& 175.15 & 48.26 & 0   \\ \hline
	10	& 97.73  & 23.84 & 0   \\ \hline
	15  & 95.00  & 25.00 & 0   \\ \hline
	20  & 70.66  & 4.80  & 0   \\ \hline
	25  & 51.60  & 3.30  & 0   \\ \hline
	30  & 49.37  & 2.32  & 0   \\ \hline
    \end{tabular}}


\end{table}

\subsection{Scalability}

Table~\ref{tab:results-agents} shows the scalability of cost adaptation to varying number of agents. The communication delay was fixed at zero and trials were simulated $100$ times each after adaptation. As more agents are added, we see search times decrease as expected, without any collisions. However, at around $30$ agents, we find that for our particular grid test case, performance gains begin to level off. This is expected, as the grid is rectangular of size $25$, a line of $25$ agents can reach maximum efficiency by sweeping across the grid (as they do).



\section{Discussion}

We demonstrate a meta-learning method of cost adaptation which leverages current swarm performance to generate an optimization problem for D-RHC. 
We demonstrate that cost adaptation successfully conforms to new environment conditions. We build a simulator on top of Unity3D, evaluate our system on a coordinated exploration task, and release all code to the public.

The benefits and applications of our method to any robotic system are clear. The adaptive nature of our formulation allows for heterogeneous mixtures of agents with different capabilities and goals to participate in the swarm. This has the potential to be used in robot convoying~\cite{shkurti2017aqua}, cooperative exploration~\cite{trevai2003cooperative}, or heterogeneous coordination for marine monitoring~\cite{shkurti2012multi}.
Our methodology follows recent trends in meta-learning~\cite{fan2018learning} and can be built upon for similar approaches in control settings with varying degrees of prior knowledge incorporated (e.g. a neural network approximator could model the D-RHC objective with no human priors).
This work is easily expandable and a base for future implementations of adaptable systems. The heuristic-based cost adaptation methods here can leverage reinforcement learning in the future for more efficient online learning. The robustness of the methods we present here are building blocks for future advances in swarm behaviour and control systems.


\bibliographystyle{IEEEtran}
\bibliography{references}

\end{document}